\documentclass{article}

    \PassOptionsToPackage{numbers, compress}{natbib}
   \usepackage[final]{iai_neurips_2024}

\usepackage[utf8]{inputenc} % allow utf-8 input
\usepackage[T1]{fontenc}    % use 8-bit T1 fonts
\usepackage{hyperref}       % hyperlinks
\usepackage{url}            % simple URL typesetting
\usepackage{booktabs}       % professional-quality tables
\usepackage{amsfonts}       % blackboard math symbols
\usepackage{nicefrac}       % compact symbols for 1/2, etc.
\usepackage{microtype}      % microtypography
\usepackage{xcolor}         % colors
\usepackage{graphicx}
\usepackage{subcaption}
\usepackage{booktabs}
\usepackage{comment}
\title{Aligning Characteristic Descriptors with Images for Human-Expert-like Explainability}

\author{
 Bharat Chandra Yalavarthi \\
 University at Buffalo\\
  Buffalo, NY  \\
  \texttt{byalavar@buffalo.edu} \\
  \And
  Nalini Ratha \\
  University at Buffalo \\
  Buffalo, NY \\
  \texttt{nratha@buffalo.edu} \\
}
\begin{document}

\maketitle

\begin{abstract}
In mission-critical domains such as law enforcement and medical diagnosis, the ability to explain and interpret the outputs of deep learning models is crucial for ensuring user trust and supporting informed decision-making. Despite advancements in explainability, existing methods often fall short in providing explanations that mirror the depth and clarity of those given by human experts. Such expert-level explanations are essential for the dependable application of deep learning models in law enforcement and medical contexts. Additionally, we recognize that most explanations in real-world scenarios are communicated primarily through natural language. Addressing these needs, we propose a novel approach that utilizes characteristic descriptors to explain model decisions by identifying their presence in images, thereby generating expert-like explanations. Our method incorporates a concept bottleneck layer within the model architecture, which calculates the similarity between image and descriptor encodings to deliver inherent and faithful explanations. Through experiments in face recognition and chest X-ray diagnosis, we demonstrate that our approach offers a significant contrast over existing techniques, which are often limited to the use of saliency maps. We believe our approach represents a significant step toward making deep learning systems more accountable, transparent, and trustworthy in the critical domains of face recognition and medical diagnosis.
\end{abstract}

\section{Introduction}

%Deep learning models have revolutionized many applications, significantly improving accuracy, efficiency, and scalability. However, they lack transparency, leaving users unaware of the rationale behind their decisions. This poses a serious problem for security applications, such as face recognition systems, which require accountability and transparency to identify biases or failures \cite{xfr}. In legal scenarios, identity search decisions must be justified, similar to how facial forensic examiners testify in court \cite{fourP}. Investigations into commercial face recognition systems have revealed inherent biases, with a notable incident involving the wrongful arrest of a woman due to a flawed match \cite{nyt} \cite{nyt2}. Similarly, in computer-aided diagnosis, the black-box nature of these models makes it difficult for physicians to trust and explain the decisions made by deep learning models. Furthermore, given the critical nature of the medical diagnosis, it is essential to ensure that models are unbiased and free of any spurious correlations which are challenging with black-box models \cite{cliplung}. The incorporation of interpretability or explainability mechanisms holds promise in mitigating such challenges by providing the reasoning behind the decisions, facilitating effective debugging processes, and illuminating the underlying biases ingrained within the model \cite{cce}.

Deep learning models have revolutionized various applications, enhancing accuracy, efficiency, and scalability. However, their lack of transparency poses significant risks in security applications like face recognition, where accountability is crucial to identify biases or failures \cite{xfr}. In legal contexts, identity search decisions must be justified, similar to how facial forensic examiners testify in court \cite{fourP}. Investigations have exposed biases in commercial face recognition systems, exemplified by a wrongful arrest due to a flawed match \cite{nyt} \cite{nyt2}. In computer-aided diagnosis, the black-box nature of these models makes it difficult for physicians to trust and explain decisions, raising concerns about biases and spurious correlations \cite{cliplung}. Incorporating explainability or interpretability mechanisms can address these challenges by clarifying decision-making processes, aiding in debugging, and revealing model biases \cite{cce}

\begin{figure}
  \centering{\includegraphics[width=0.97\textwidth]{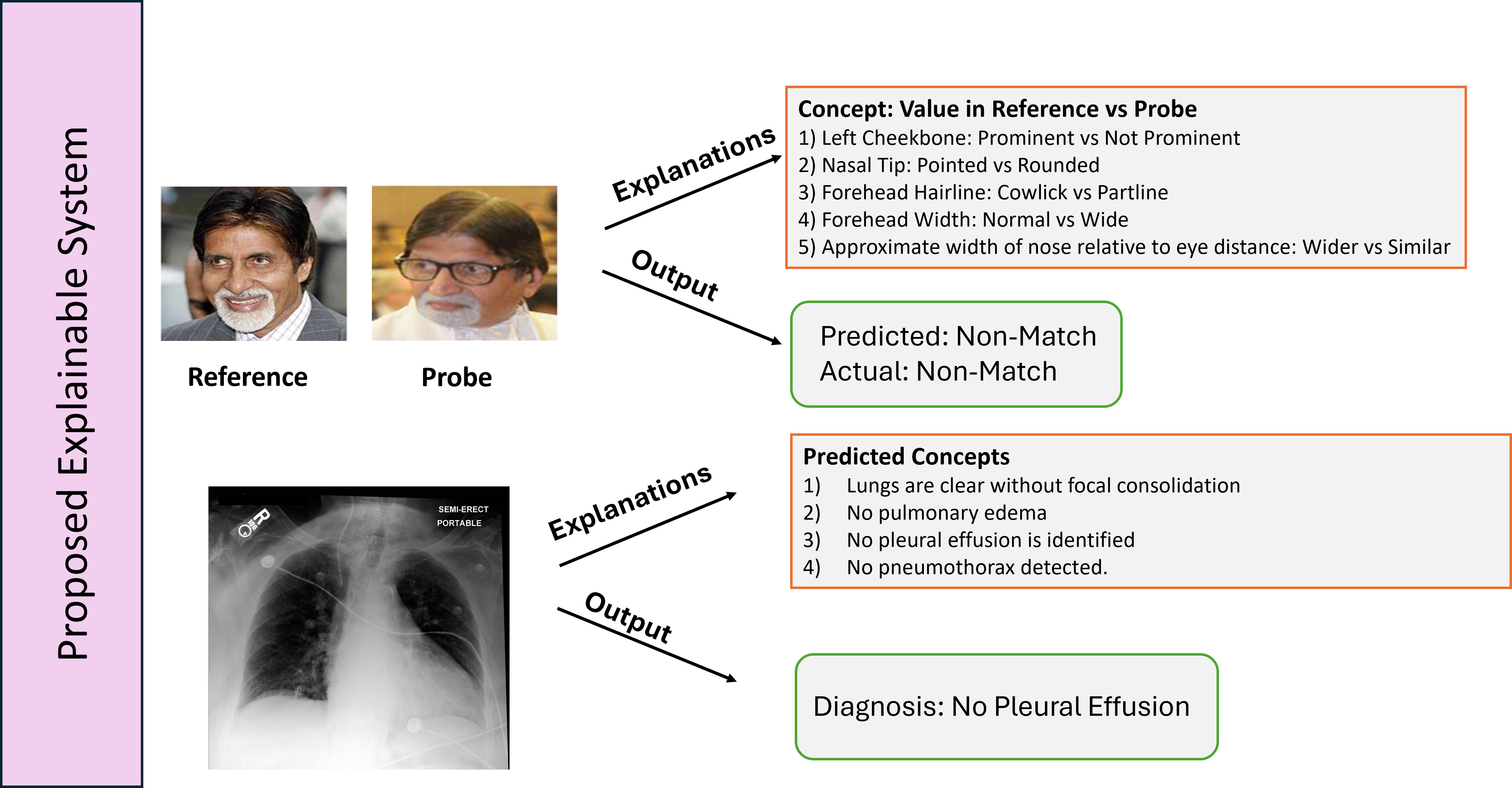}}
\caption{Illustration of our proposed explainable system for face recognition and chest x-ray diagnosis.}
\label{fig:overview}
\end{figure}

Although there are several existing methods relating to explainable face recognition \cite{xfr, xcos, comp, agno} they focus on indistinct visual explanations which are less interpretable than textual explanations \cite{fourP}. Visual explanations can lack the nuanced detail and context often provided in textual explanations and may be subject to misinterpretation \cite{fourP}. %Moreover, the existing method \cite{sim2word} that offers textual explanations necessitates the availability of labeled face attributes. It also presents practical limitations by constraining the scope of explanation solely to a few labeled face attributes and lacking the ability to provide explanations similar to face forensic experts.  
While prior works dealt with explaining the chest x-ray diagnosis made by deep learning models, none have explored using fine-grained, and atomic characteristic descriptors for providing radiologist-like explanations. The prior work can be mainly categorized into the explainable x-ray report generation \cite{inter} \cite{promptmrg}, or saliency maps for the x-ray images to depict the important regions for the diagnosis \cite{iai}. The ability of deep learning systems to provide human-expert-like explanations can help in regulatory compliance especially in legal proceedings, improved communication, and increased user trust and acceptance \cite{fourP}. Contrary to the prior work, our approach employs precisely defined characteristic descriptors which are textual concepts to emulate the explanatory abilities of human experts in justifying facial recognition and chest x-ray diagnosis decisions, while also ensuring the faithfulness of the explanations through self-explainable architecture. An example of the explanations provided by our framework is given in figure \ref{fig:overview}. 

%\begin{figure*}
%\centerline{\includegraphics[width=0.95\textwidth]{figures/overview.png}}
%\caption{Illustration of our proposed explainable system for face recognition %and chest x-ray diagnosis.}
%\label{fig:overview}
%\end{figure*}

Our framework is designed to function in both supervised and unsupervised environments and we demonstrate the efficacy of our framework in both contexts. For the unsupervised case, we utilize face recognition, and for the supervised case, we apply it to chest x-ray diagnosis.  
The characteristic descriptors or concepts (used interchangeably) for face recognition are derived from the facial features standard published by FISWG \cite{fiswg} for morphological analysis in face comparison for face recognition. For x-ray diagnosis, we extracted the descriptors from the radiology reports of the MIMIC-CXR  \cite{mimic} dataset. Our key contribution is proposing a generic explainable framework capable of functioning in both supervised and unsupervised contexts, which can be used to provide expert-like explanations to any classification decisions made by deep learning models. We show the results of our method which provides a coherent, faithful, user-friendly, and expert-like textual explanation for the tasks of face recognition, and x-ray diagnosis.  
%    \item We demonstrate the debugging and analysis capabilities of our approach using counterfactual examples in face recognition, identifying several underlying model biases

    %\item We show through our results on five standard face recognition datasets that we can devise self-explainable models for face recognition without compromising performance.     
%\end{itemize}

\section{Related Work}

The prior work related to our proposed approach can be broadly divided into three categories i) Vision Language Models (VLMs) for Explainability ii) Explainable Face Recognition and iii) Explainable x-ray diagnosis 

\textbf{VLMs for Explainability:} Pre-trained VLMs like CLIP \cite{clip}, and ALIGN \cite{align} have shown good performance in in various tasks across several domains in zero-shot and fine-tuning settings \cite{robust, adapter}. %In a zero-shot setting, the similarity between a textual description of the class labels and an image is used for classification. Various fine-tuning techniques also exist including training a linear probe with residual connection\cite{adapter}, tuning the image encoder or textual embeddings, weight ensembling of zero-shot and fine-tuned versions \cite{robust}. 
Recently there have been several works where VLM's like CLIP were used to design explainable image classification models. \cite{bottle, labelFree, visualLLM} uses the concept bottleneck layer formed by aligning textual concepts and images for explainable image classification. Candidate concepts are usually generated by prompting the LLMs and undergo a selection process. We extend the use of VLMs for explainability to provide expert-level explanations using characteristic descriptors.

\textbf{Explainable Face Recognition:} Most prior work (examples in supplementary material) focuses on generating visual explanations in the form of saliency maps, highlighting the regions of the face that the model considers when making a decision. \cite{blackbox1, comp, blackbox2, xfr, peri, agno,backprop, xcos} identify these important regions through  different techniques including occlusion, perturbation, similarity measurement, and attention. %forms of occlusion or perturbation and how they affect the face-matching decision. Another approach in creating these saliency maps \cite{} \cite{} is identifying regions of the face pairs that are similar and dissimilar or identifying regions that lead to an imposter decision. \cite{} provides both patch-wise similarity of face images and attention weights indicating the importance of each patch in making a matching decision.    
Visual explanations have several disadvantages, they are not precise, fine-grained, and may be subject to interpretation \cite{fourP}. In human communications, an explanation response is usually in a textual medium either in spoken or written form as this can provide clear and concise explanations in most cases \cite{fourP}. 
Unlike above works, \cite{sim2word} offers both textual and visual explanations by training separate networks to identify face attributes and using counterfactual examples to determine key attributes. However, it relies on a limited set of constrained attributes, reducing explanation precision, depth and coverage compared to our method. \cite{chatGPT1} \cite{chatGPT2} evaluate the usage of ChatGPT for performing and explaining face recognition. But this method can have reliability issues as these models can hallucinate and faithfulness of the explanations can't be ensured. Moreover, unlike ours, \cite{sim2word} \cite{chatGPT1} \cite{chatGPT2} do not provide explanations similar to that of a forensic expert justifying a face match decision. %Our approach provides explanations through textual characteristic descriptions of facial features used by face forensic experts with a self-explainable architecture to ensure faithfulness. 

\textbf{Chest X-ray Diagnosis:} Prior works like \cite{cliplung} and \cite{medklip}, employ saliency maps and report features to enhance disease classification, while others, such as \cite{iai}, model radiologist gaze for explanation. Methods like \cite{inter} and \cite{control} link image regions to report sections and utilize longitudinal data for interpretable report generation, respectively. Additionally, \cite{lmimic} leverages model-produced explanations to improve classification, and \cite{multimodal} addresses explaining image-text alignment in vision-language models. In contrast, our approach provides radiologist-like explanations using radiologist-defined descriptors for diagnostic decisions.

\section{Methodology}

\begin{figure*}[htbp]
\centerline{\includegraphics[width=0.90\textwidth]{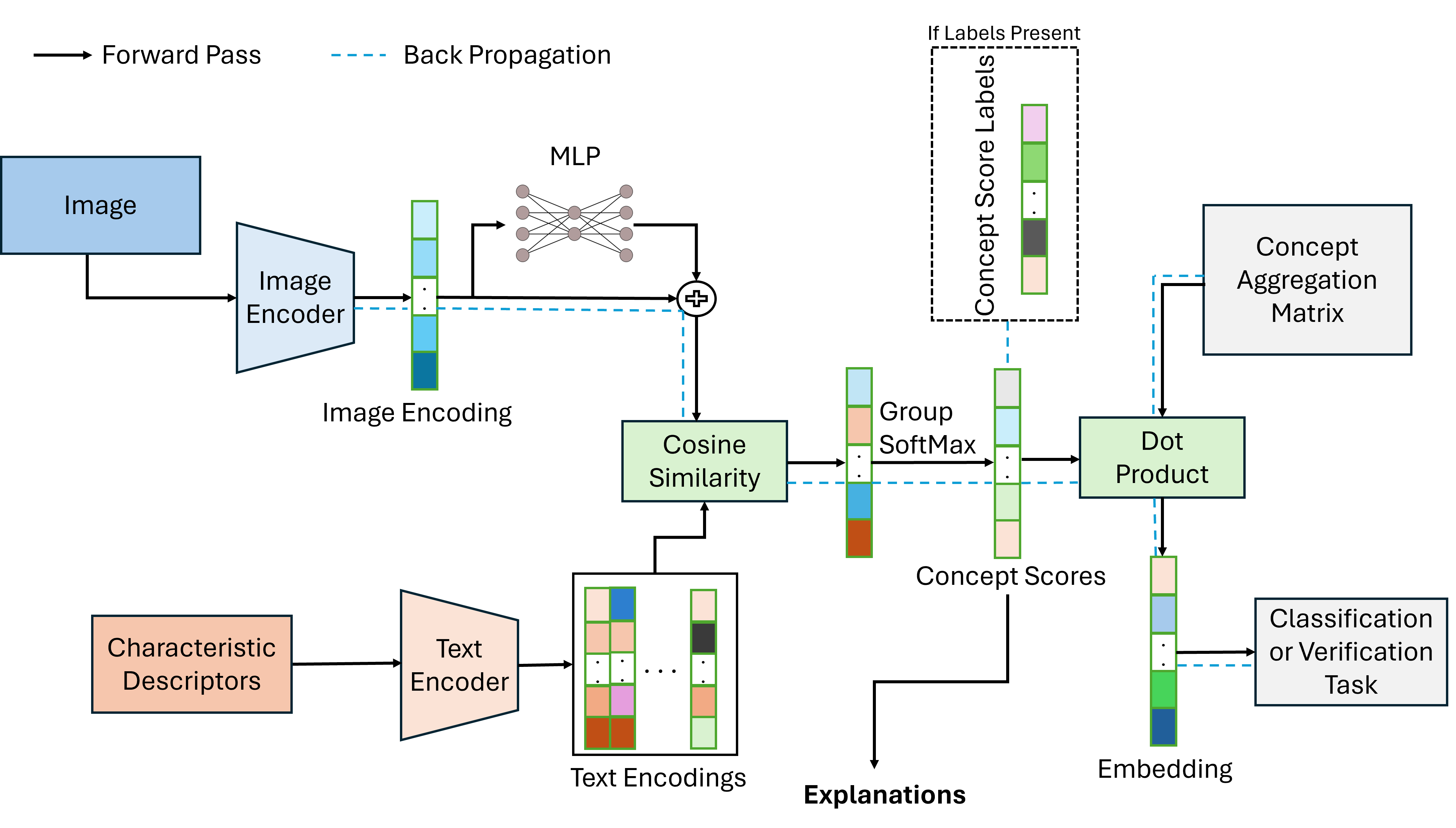}}
\caption{Proposed architecture produces explanations and embeddings for tasks like classification or face recognition. It can work with or without concept supervision. If concept  labels are available, they are used for back-propagating the loss; otherwise, only the task loss is back-propagated.}
\label{fig:arch}
\end{figure*}

Our proposed novel explainable methodology is based on using characteristic descriptors for providing human-expert like explanations. %For face recognition, these characteristic descriptor are derived from the facial features standard published by Facial Identification Scientific Working Group (FISWG) \cite{fiswg} for purpose of morphological analysis in face comparison and recognition by forensic experts. For chest X-Ray diagnosis, they were taken from MIMIC-CXR radiology reports \cite{mimic}.
Based on such characteristic descriptors described by human experts we create a set of textual concepts denoted by $C =\{c_1,c_2,..c_n\}$. These concepts are divided into groups based on the component they are describing.
Inspired by the concept bottleneck models (CBM) \cite{cbm} to design inherently explainable models we use a bottleneck layer in our proposed methodology and employ CLIP \cite{clip} \cite{openClip} to identify these textual concepts in the images. The similarity score between the textual and image embeddings produced by CLIP can give us the concept scores that form the bottleneck layer in our architecture. The concept scores of each image reveal the extent of a concept's presence and are used for providing explanations. Let $X \in R^{H*W*D}$ denote an image where $H$, $W$, $D$ are the height, width, and channels of the image, and $y$ its label. We denote the image encoder module of the CLIP as $E_i$ and the text encoder module as $E_t$. The dot product between the encodings of $E_i$ and $E_t$ shows the match between the image and the text modalities. To fine-tune CLIP we use a skip connection to extract image embeddings as first detailed in \cite{adapter}. Image encoding $I \in R^d$ is extracted using the following equation:

\begin{equation}
    I = \alpha * E_i(X) + (1-\alpha) * F_i(E_i(X))
\end{equation}

where $F_i$ denotes a two-layer network with ReLU activation, which down-samples and then up-samples the embedding back to its original size.

On the other hand, each of the tokenized concepts are fed into $E_t$ to get the text encodings $T \in R^{N*d}$. We compute the cosine similarity between $I$ and and each encoding in $T$ to get concept scores $S \in R^{N}$which represents the presence of concepts in an image. To better represent these concept scores and the dependencies within a concept group we apply SoftMax independently within each group (denoted as Group SoftMax) of the concept set to obtain $S_{sm}$.  Group SoftMax enhances the representation of embeddings by emphasizing the most activated concept within each group, thereby improving both accuracy and explainability, as demonstrated in an ablation study in the supplementary material. We then transform the $S_{sm}$ using a learned concept aggregation matrix $W \in R^{N*m}$ to get the final embedding $X_{emb} \in R^m$ used for downstream tasks like face recognition or disease classification. The concept aggregation matrix can be further used to understand the feature weights of the concepts. Figure \ref{fig:arch} shows the architecture of the proposed explainable framework. 

\begin{figure*}[htbp]
\centering
    \begin{subfigure}[b]{0.45\textwidth}
        \centering
        \includegraphics[width=\textwidth]{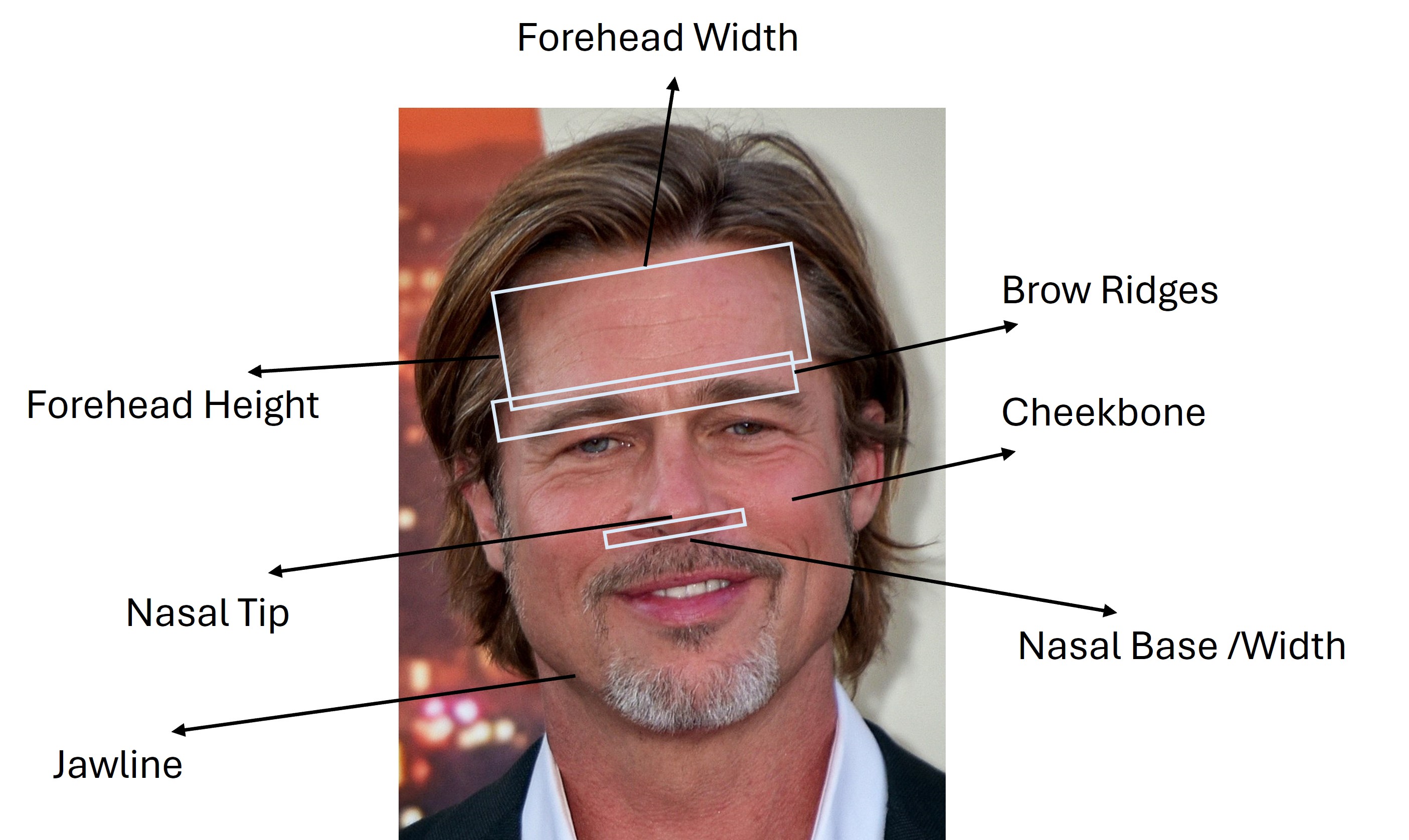}
        \caption{}
        \label{fig:comp1}
    \end{subfigure}
    \hfill
    \begin{subfigure}[b]{0.50\textwidth}
        \centering
        \includegraphics[width=\textwidth]{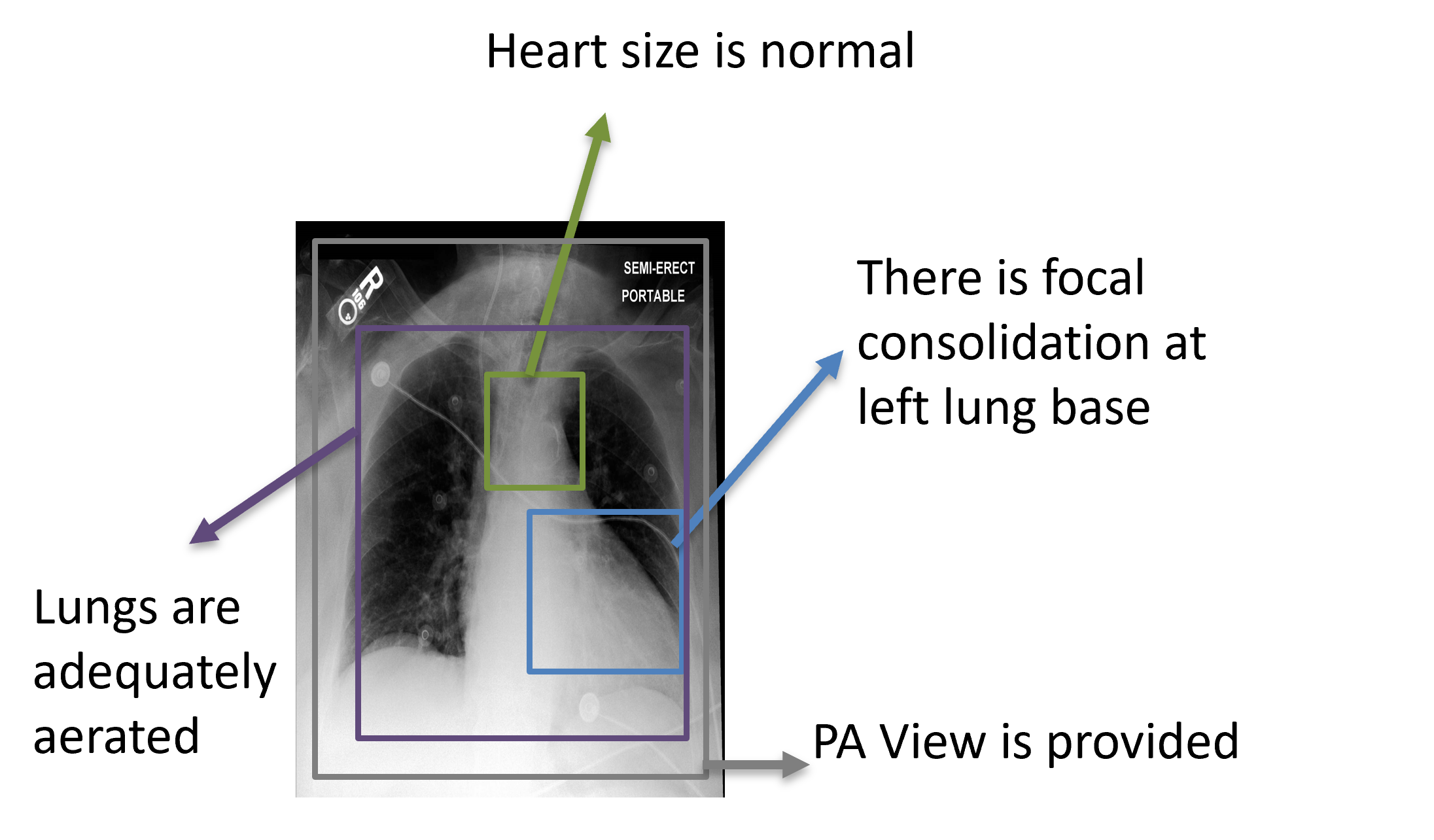}
        \caption{}
        \label{fig:comp2}
    \end{subfigure}
\caption{Illustrative examples of some of the a) facial characteristic descriptors as recommended in the FISWG Facial Image Comparison Guide.  b) characteristic descriptors for chest x-ray diagnosis as defined by radiologists.}
\label{fig:comp}
\end{figure*}

\subsection{Face Recognition}
The FISWG guide listing the facial features to be used for face comparison and morphological analysis is commonly followed by forensic experts. It has detailed characteristic descriptors for each of the nineteen components of the human face. Figure \ref{fig:comp} (a) shows some examples of the characteristic descriptors presented in FISWG guide \cite{fiswg}. We use 120 such descriptors from the guide in our explainable face recognition system. Given a reference and probe face images, we get the face embeddings of these by passing it through the proposed architecture and compute if it is a match or not based on their cosine similarity (shown in figure \ref{fig:faceRecog} of supplementary material). Since no face dataset includes labeled characteristic descriptors for faces, and most face descriptors are general features fairly detectable by a generic model, we rely on pre-trained image and text encoders of CLIP\cite{openClip} trained on LAION 2B internet data to identify these descriptors in an unsupervised manner. We further fine tuned our model for face recognition using the quality adaptive margin loss function proposed in AdaFace \cite{adaface}. We tune only the fully connected layer towards the end of the image encoder. Additionally, we adopted the idea from \cite{adapter} to add a fully connected layer on top of the image encoder and form a residual connection between their respective outputs (denoted as Adaptive FT). The ablation studies involved in determining the optimal architecture are described in the supplementary material.

%\begin{figure}[htbp]
%\centering
%\includegraphics[width=0.53\textwidth]{figures/CFEx.png}
%\caption{Example of counterfactual explanation. Modifying the %concepts shown corrects the model decision.}
%\label{fig:cfEx}
%\end{figure}

\subsection{Chest X-Ray Diagnosis}

Features in chest x-ray images are more niche and subtle, making it challenging for a generic CLIP model to capture them. We use a CLIP model pre-trained on chest X-Ray images and radiologist reports which is proposed by \cite{cxrClip} as our image and text encoders. The CLIP image encoder and the text encoder are frozen while only the fully connected layers are trained. As we can extract the concept labels from radiologist reports, we can supervise the concepts unlike in face recognition where there was no supervision. Given a corpus of radiology reports, we extract the atomic, fine-grained characteristic descriptors from them using the Mistral 7B language model. We prompt it to disentangle the descriptors to separate sentences from the report (shown in figure \ref{fig:cdextract} of supplementary material). Figure \ref{fig:comp} (b) shows the illustration of some of the characteristic descriptors for chest x-ray used by radiologists. 

Given a chest x-ray, its corresponding concept labels, and its diagnosis label we calculate the cosine similarity between the x-ray image and the characteristic descriptor embeddings to obtain concept scores. For obtaining concept score labels we set the concepts present in the report to max value in the calculated concept scores. L1 loss calculated between the concept scores and concept scores label is used for supervising the model to give appropriate explanations. The concept scores are further passed through a concept aggregation layer to get the logits used for making the diagnosis prediction. As the standard, we use the cross-entropy loss for classification (entire pipeline is shown in figure \ref{fig:xRayPipeline} of supplementary material).

\section{Experiments and Results}
%In this section, we present the classification/recognition performance, explanation results, and training and dataset details. Since there is no baseline in prior work that provides human-expert-like explanations using characteristic descriptors, we only compare the performance of our framework with its black-box counterparts.  
\subsection{Face Recognition}
%We evaluate our approach on several standard face recognition datasets, generate explanations for predictions made by the model. %Out of multiple open-source variants of CLIP, we chose the model with ViT-L/14 architecture trained on the LAION-2B dataset as it was the best-performing variant for face recognition in our experiments shown in supplementary material %\ref{tab:clip}.  

\textbf{Datasets and Training:}
A random subset of 500k images from MS1MV2 \cite{arcface} dataset containing 5.8M images with 85K identities was used for fine-tuning our model.
Standard face recognition datasets including LFW \cite{lfw}, CFP-FP \cite{cfp}, AgeDB \cite{agedb}, CPLFW \cite{cplfw}, CALFW \cite{calfw}  are used for validation.  We resize the cropped and aligned MS1MV2 images to 224 x 224 for compatibility with CLIP's image encoder. The model is fine-tuned for 5 epochs using the AdamW optimizer with a learning rate of 0.0003, following the approach in \cite{robust}. We use the same hyper-parameter values for margin $m$ and image quality indicator concentration $h$ as in Adaface \cite{adaface}.
Table \ref{tab:perf} shows the 1:1 verification performance of our proposed approach on these five datasets compared to the black box SOTA model AdaFace \cite{adaface}. Although we lose some accuracy compared to the black box model, our approach provides explanations leading to insights and debugging capabilities, which are as important as performance in certain critical applications. 

\begin{figure*}[htbp]
\centerline{\includegraphics[width=0.95\textwidth]{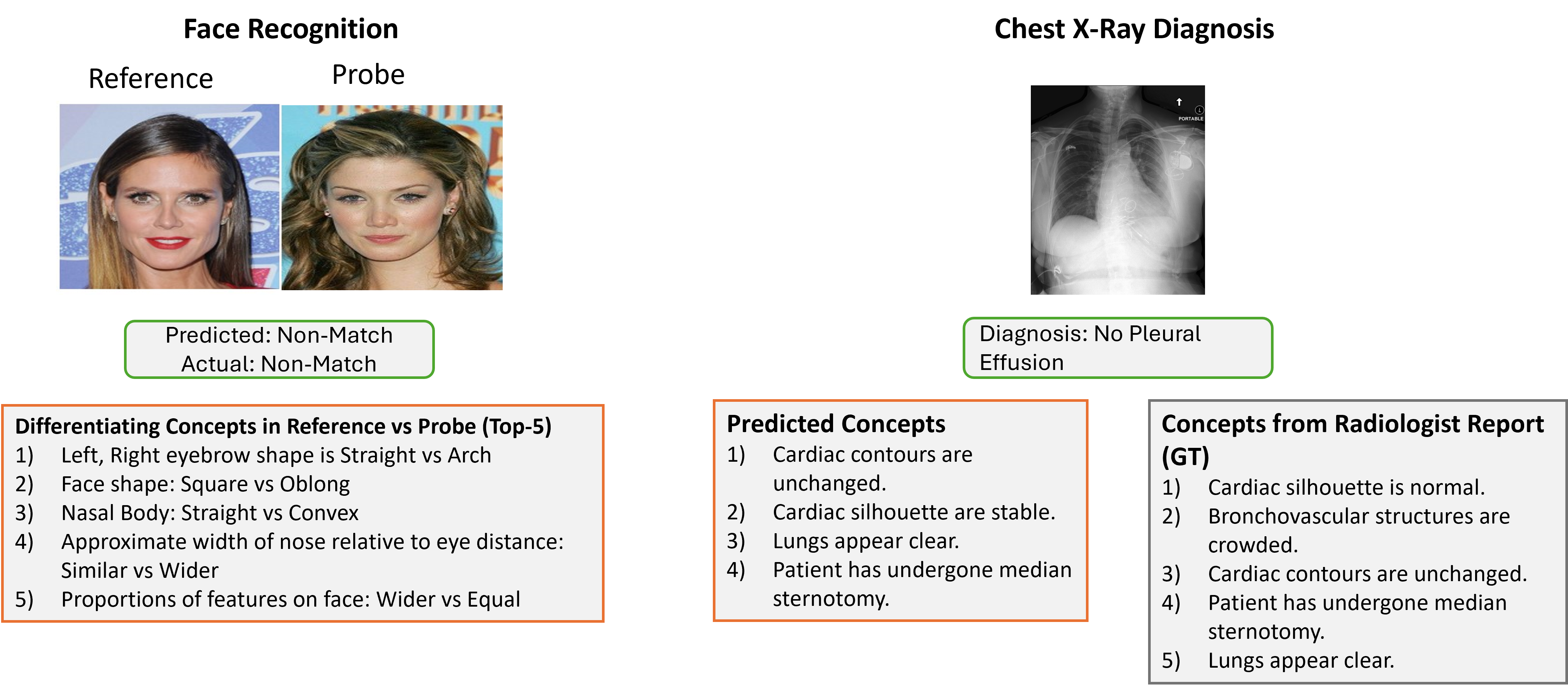}}
%\centerline{\includegraphics[width=0.85\textwidth]{figures/just2.jpg}}
\caption{Examples of explanations provided by our model for its decisions in face recognition and chest x-ray diagnosis.}
\label{fig:justification}
\end{figure*}

%TODO : Baseline
\textbf{Explanations:} We show the ability of our approach to produce faithful explanations justifying its face-matching decisions. Figure \ref{fig:justification} shows the examples of the model justifying its decision based on important concepts. The top closest concepts of reference and probe which are commonly activated in each concept group are shown for matching cases, while the top concept groups where the activated concepts differ the most between the reference and probe are shown for non-match cases.

%\begin{figure*}[htbp]
%\centerline{\includegraphics[width=0.85\textwidth]{figures/ablArch.png}}
%\caption{Average Accuracy on the five validation datasets with the experimented architecture choices.}
%\label{fig:ablArch}
%\end{figure*}

\subsection{Chest X-ray Diagnosis}

\textbf{Dataset and Training:} We use the MIMIC-CXR dataset which has chest x-rays, corresponding radiologist reports, and disease labels. For this work, we classify the presence of Pleural Effusion condition from x-ray images using a subset of around 20K samples. We make an 85\% train and 15\% test split from the chosen subset of the MIMIC-CXR data. We used the pre-trained CLIP model \cite{cxrClip} as a baseline and compared its classification performance in detecting pleural effusion against our proposed explainable architecture. We observe from table \ref{tab:perfXray} that the classification performance of our proposed architecture is similar to that of the black box while our approach has an added advantage of providing explanations. The model was fine-tuned for 10 epochs using Adam optimizer with a learning rate of 0.005 with 1:2 proportions of loss signals from classification and concept prediction.  

\textbf{Explanations:}  We do a quantitative evaluation of the explanations provided by our model with the labels extracted from radiologist reports. We use two metrics for our evaluation, i) METEOR ii) Rouge-L score for measuring the quality of explanations when compared against the ground truth. The evaluation results in table \ref{tab:expResultXRay} show that we achieve a good Rouge-L score exhibiting the fidelity of our explanations. One possible reason for the METEOR score not being higher could be the difference in the ordering of concepts between the predictions and the ground truth. We do not optimize for ordering as we treat all ground truth concepts as equally important.  
Figure \ref{fig:justification} shows example results of our explainable chest x-ray diagnosis system.

%\begin{figure*}[htbp]
%\centerline{\includegraphics[width=0.95\textwidth]%{figures/example.png}}
%\caption{Examples of the explanations produced by the model for chest X-ray diagnosis.}
%\label{fig:xRayRes}
%\end{figure*}

\begin{table}[!htbp]
  \centering
  \begin{minipage}[t]{0.45\textwidth}
    \centering
    \caption{Performance (1:1 verification accuracy) of our proposed explainable face recognition approach on 5 benchmark datasets compared with black box SOTA \cite{adaface}.}
    \label{tab:perf}
    \renewcommand{\arraystretch}{1}
    \setlength{\tabcolsep}{5pt}
    \begin{tabular}{p{2cm}p{1.6cm}p{1.8cm}} % Adjust the width as needed
      \toprule
      \textbf{Dataset} & \textbf{Proposed (Explainable)} & \textbf{Baseline (Black box - SOTA) } \\
      \midrule
      LFW \cite{lfw}     & 98.38\% & 99.82\% \\
      CFP-FP \cite{cfp}  & 90.88\% & 98.49\% \\
      CALFW \cite{calfw} & 89.36\% & 96.08\% \\
      AgeDB \cite{agedb} & 81.18\% & 98.05\% \\
      CPLFW \cite{cplfw} & 87.71\% & 93.13\% \\
      \midrule
      Average            & 89.50\% & 97.11\% \\
      \bottomrule
    \end{tabular}
  \end{minipage}%
  \hspace{0.04\textwidth} % Adjust the space between the tables as needed
  \begin{minipage}[t]{0.45\textwidth}
    \centering
    \caption{Performance of our proposed explainable chest x-ray diagnosis approach on MIMIC-CXR dataset compared with black box model.}
    \label{tab:perfXray}
    \begin{tabular}{p{2cm}p{2cm}p{2cm}} % Adjust the alignment of each column as needed
      \toprule
      \textbf{Metric} & \textbf{Proposed (Explainable)} & \textbf{Baseline (Black Box)} \\
      \midrule
      Accuracy & 83.78\% & 86.64\% \\
      Precision & 84.77\% & 86.4\% \\
      Recall & 83.78\% & 87.3\% \\
      F1-Score & 84.19\% & 86.8\% \\
      \bottomrule
    \end{tabular}
  \end{minipage}
\end{table}

\begin{comment}
\begin{table}[!htbp]
  \centering
  \caption{Performance of our proposed explainable chest x-ray diagnosis approach on MIMIC-CXR dataset compared with black box model.}
  \label{tab:perfXray}
  \begin{tabular}{p{2cm}p{2cm}p{2.2cm}} % Adjust the alignment of each column as needed
    \toprule
    \textbf{Metric} & \textbf{Proposed (Explainable)} & \textbf{Baseline (Black Box) } \\
    \midrule
    Accuracy & 83.78\% & 86.64\% \\
    Precision & 84.77\% & 86.4\% \\
    Recall & 83.78\% & 87.3\% \\
    F1-Score & 84.19\% & 86.8\% \\
    \bottomrule
  \end{tabular}
\end{table}
\end{comment}

\begin{table}[!htbp]
  \centering
  \caption{Text evaluation metrics used for comparing the explanations produced by model for chest x-ray diagnosis with ground truth.}
  \label{tab:expResultXRay}
  \begin{tabular}{ll}
    \toprule
    \textbf{ROUGE-L} & \textbf{METEOR} \\
    \midrule
    0.31  & 0.27 \\
    \bottomrule
  \end{tabular}
\end{table}
%\begin{comment}

\section{Conclusion}
In this work, we propose a methodology to explain the decisions made by deep learning systems similar to that of a human expert thereby providing improved explainability and expert level insight. We show that we can design models that give faithful and concrete explanations like a human expert using characteristic descriptors.  Through our experiments, we show the performance on benchmark datasets proving the efficacy of our model in providing explanations without significantly affecting the performance in comparison with black-box models. We hope our proposed method with the ability to produce consumable and verifiable descriptions can address transparency and trustworthiness in face recognition and x-ray diagnosis systems.

\bibliographystyle{splncs04}
\bibliography{Styles/iai_neurips_2024}

%\section*{References}

%%%%%%%%%%%%%%%%%%%%%%%%%%%%%%%%%%%%%%%%%%%%%%%%%%%%%%%%%%%%

\appendix

\section{Appendix / supplemental material}

\subsection{Explainable Face Recognition}

\begin{figure*}[htbp]
\centerline{\includegraphics[width=0.95\textwidth]{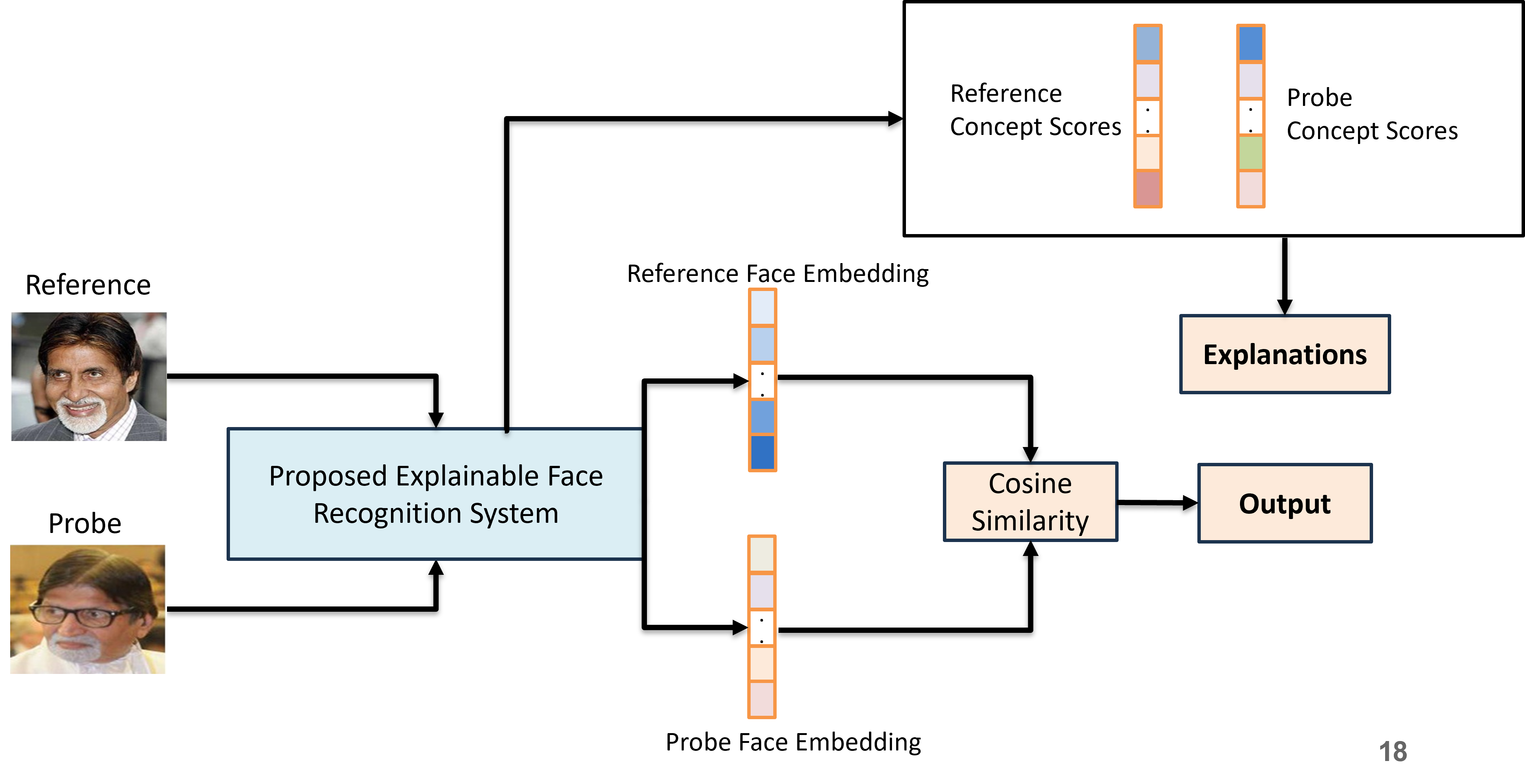}}
\caption{Proposed Explainable Face Verification Pipeline.}
\label{fig:faceRecog}
\end{figure*}

\subsubsection{Ablation Study}
\textbf{CLIP Model: } As there are multiple variants of CLIP differing in the architecture and training dataset we evaluate various CLIP variants based on their Zero-shot performance in 1:1 verification task. We report the average accuracy on the five validation datasets of each variant in table \ref{tab:clip}. Based on these results we chose ViT-L/14 trained on LAION-2B dataset variant for all further experiments. Interestingly, it outperformed other variants with larger parameters such as VIT-H-14-quickgelu and VIT-H-14-378-quickgelu, and also a variant with the same architecture trained on a different dataset.

\begin{table*}%[!htbp]
  \centering
  \caption{CLIP Variants and Zero-Shot Accuracy on Face Verification. \newline}
  \begin{tabular}{p{4cm}p{3cm}p{3cm}} % Adjust the width of each column as needed
    \toprule
    \textbf{Variant} & \textbf{Training Dataset} & \textbf{Accuracy} \\
    \midrule
    ViT-B/16 & DataComp-1B & 68.72\%    \\
    
     ViT-L/14 & OpenAI's WIT & 71.21\%  \\
    
   ViT-H-14-378-quickgelu & dfn5b & 72.29\% \\
  
   ViT-H-14-quickgelu & dfn5b & 72.87\% \\
  
    \textbf{ViT-L/14} &  \textbf{LAION-2B} & \textbf{74.11\%}  \\
   \bottomrule
  \end{tabular}
  \label{tab:clip}
\end{table*}

\textbf{Architecture:} In our experiments with various architectural configurations, we discovered that implementing group SoftMax improved verification accuracy. Furthermore, the combination of CLIP MLP unfreezing, group SoftMax, and a linear layer yielded the highest performance for face recognition tasks among the configurations tested as shown in table \ref{tab:abl}. In the nest experiment we experiment with various fine-tuning methods.   

\begin{table*}[!htbp]
  \centering
  \caption{Ablation Study of Architecture Choices for Fine-Tuning. \newline}
  \begin{tabular}{p{9cm}p{3cm}} % Adjust the width of each column as needed
    \toprule
    \textbf{Architecture} & \textbf{Face Recognition Accuracy}  \\
    \midrule
    CLIP Zero-Shot (No FT) & 74.11\%  \\
  
    CLIP Zero-Shot (No FT) + Group SoftMax &  76.91\%\\
  
    CLIP Zero-Shot (No FT) + Group SoftMax + Linear Layer & 76.91\%\\

     CLIP Image-Encoder MLP FT + Group SoftMax + Linear Layer \textbf{(Base)} & 83.72\%\\
     %\hline
     % \textbf{Base + Adaptive FT}  & \textbf{89.50\%}\\
     \bottomrule

  \end{tabular}
  \label{tab:abl}
\end{table*}

\textbf{Fine-Tuning:}
To ensure explainability, it is crucial to fine-tune the CLIP backbone without compromising its text-image alignment capability, as the faithfulness and validity of the explanations are directly dependent on it. Previous work has demonstrated the strong performance of CLIP in zero-shot image classification settings due to its robust capability to align text and images. We use the zero-shot performance of our fine-tuned CLIP on CIFAR-10 as a proxy for its alignment capability to determine the appropriate level of fine-tuning. Given that strong zero-shot performance is indicative of CLIP's alignment capability, we believe this can serve as a reliable proxy. Our goal in this experiment is to fine-tune the model to enhance face recognition accuracy without affecting its alignment capability.
As shown in table \ref{tab:abl1}  fine-tuning the entire image encoder part of CLIP has led to the best face recognition accuracy of all the experimented cases, but as evident from the alignment proxy accuracy it has lost its ability to align text and images rendering it unusable for explanations. Weight ensemble fine-tuning which was proposed in \cite{robust} interpolates the weights of zero-shot and end-to-end fine-tuned models to bring the generality of zero-shot model to fine-tuned one. Although this provided improvement in the alignment but was not able to bring accuracy to the levels of the zero-shot model. Further experiments have proved that fine-tuning using Adaptive-FT as proposed in \cite{adapter} or fine-tuning the last MLP layer of the CLIP image encoder along with group SoftMax and a linear layer (denoted as Base) showed improvements in face recognition while not losing significantly on the alignment. Moreover, combining both these techniques was shown to be the best balance between face recognition performance and text-image alignment.

\begin{table*}[!htbp]
  \centering
  \caption{Ablation Study of CLIP Fine-Tuning methods. \newline}
  \begin{tabular}{p{6cm}p{3cm}p{3cm}} % Adjust the width of each column as needed
    \toprule
    \textbf{Fine-Tuned Model} & \textbf{Face Recognition Accuracy} & \textbf{Alignment Proxy Accuracy} \\
    \midrule
    CLIP Zero-Shot (No FT) & 74.11\%  & 96.91\% \\
   % \hline
    Entire Image Encoder FT & 94.80\% & 12.59\%\\
   % \hline
    Weight Ensemble FT \cite{robust} & 76.72\% & 36.16\%\\
    %\hline
     Adaptive FT \cite{adapter}   & 83.50\% & 96.82\%\\
     %\hline
      Base & 83.72\% & 93.59\%\\
     %\hline
     \textbf{Base + Adaptive FT} & \textbf{89.50\%} & \textbf{94.27\%}\\
     \bottomrule
  \end{tabular}
  \label{tab:abl1}
\end{table*}

\textbf{Face Recognition Training:}
We have experimented with existing state-of-the-art methods of AdaFace \cite{adaface}, ArcFace \cite{arcface}, and CosFace \cite{cosface},  for training face recognition models to find the best suitable one to fine-tune our model whose results are shown in table \ref{tab:abl2}. Varying slightly from the AdaFace's original parameter values we find that for our setting margin $m$=0.5, and image quality indicator concentration $h$=0.33 was the best performing combination. 

\begin{table*}[!htbp]
  \centering
  \caption{Ablation Study of our approach with various face recognition training methods.}
  \label{tab:abl2}
  \begin{tabular}{lc}
    \toprule
    \textbf{Method} & \textbf{Accuracy} \\
    \midrule
    CosFace (m=0.35) & 80.14\% \\
    ArcFace (m=0.50) & 83.24\% \\
    \textbf{AdaFace (m=0.5)} & \textbf{89.50\%} \\
    \bottomrule
  \end{tabular}
\end{table*}

\subsection{Explainable Chest X-ray Diagnosis}

\begin{figure*}[htbp]
\centerline{\includegraphics[width=0.95\textwidth]{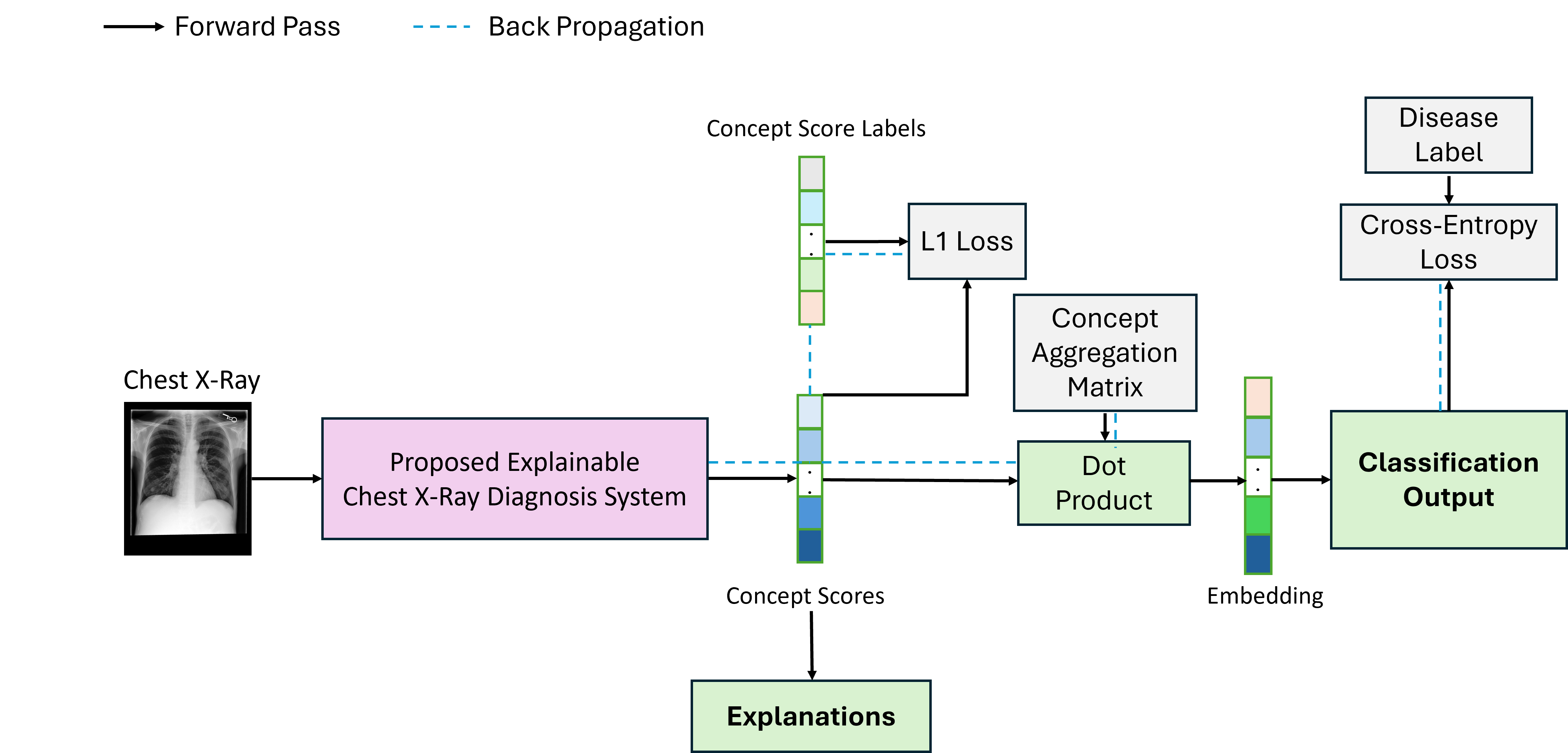}}
\caption{Proposed Explainable Chest X-Ray Diagnosis Pipeline.}
\label{fig:xRayPipeline}
\end{figure*}

\begin{figure*}[htbp]
\centerline{\includegraphics[width=0.95\textwidth]{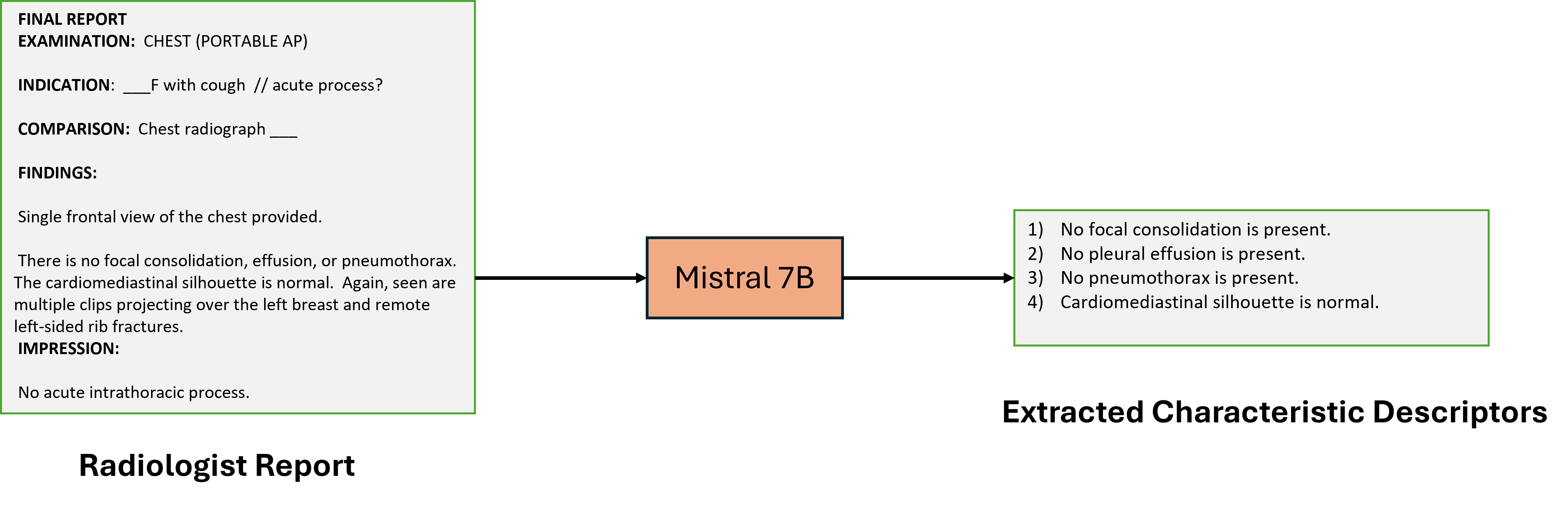}}
\caption{An example of extracting characteristic descriptors from radiologist reports. We prompt the Mistral 7B model to extract atomic concepts from the report findings.}
\label{fig:cdextract}
\end{figure*}

\subsection{Explanations provided by prior work}
\begin{figure*}[htbp]
\centerline{\includegraphics[width=0.95\textwidth]{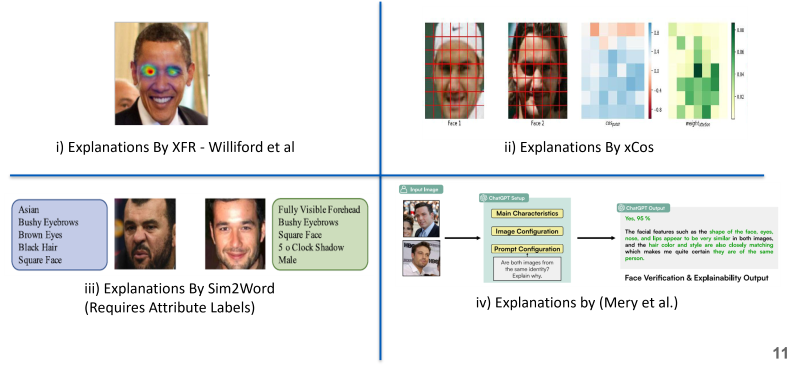}}
\caption{Existing explainable face recognition methods.}
\label{fig:existing1}
\end{figure*}

\begin{figure*}[htbp]
\centerline{\includegraphics[width=0.90\textwidth]{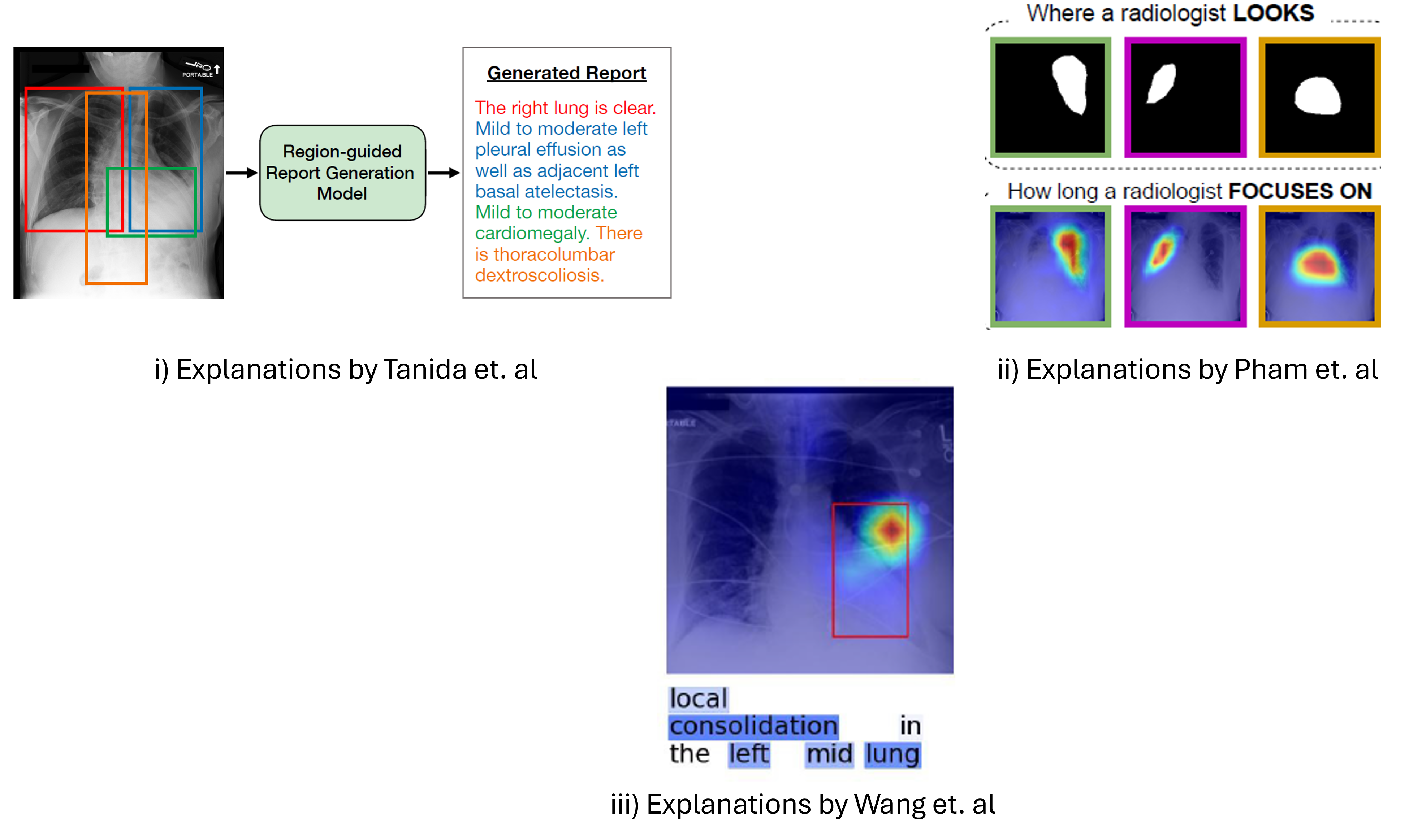}}
\caption{Existing explainable chest X-ray diagnosis methods.}
\label{fig:existing2}
\end{figure*}

Figures \ref{fig:existing1} and \ref{fig:existing2} shows the explanations provided by current explainability methods in face recognition and chest x-ray diagnosis respectively. 

%Optionally include supplemental material (complete proofs, additional experiments and plots) in appendix.
%All such materials \textbf{SHOULD be included in the main submission.}

%%%%%%%%%%%%%%%%%%%%%%%%%%%%%%%%%%%%%%%%%%%%%%%%%%%%%%%%%%%%

\end{document}